\documentclass[10pt,twocolumn,letterpaper]{article}

\usepackage[pagenumbers]{cvpr} 

\usepackage[utf8]{inputenc} 
\usepackage[T1]{fontenc}    
\usepackage{url}            
\usepackage{booktabs}       
\usepackage{amsfonts}       
\usepackage{nicefrac}       
\usepackage{microtype}      
\usepackage{graphicx}
\usepackage{colortbl}
\usepackage{array}
\usepackage{caption}
\usepackage{makecell}
\usepackage{wrapfig}
\usepackage{subcaption} 
\usepackage{graphicx}
\usepackage{natbib}
\usepackage{pifont}
\usepackage[dvipsnames]{xcolor}
\usepackage{pifont} 

\usepackage{tcolorbox}
\tcbuselibrary{skins,breakable,listings}
\usepackage{xcolor}
\usepackage{wrapfig}
\usepackage{multirow}
\usepackage{multicol}
\usepackage{listings} 
\usepackage{titletoc} 
\usepackage{fancyvrb}
\usepackage{tcolorbox} 

\usepackage[accsupp]{axessibility}
\definecolor{repa_blue}{RGB}{242, 242, 255}
\usepackage{soul}

\definecolor{jointgreen}{HTML}{257849}
\definecolor{jointbg}{HTML}{DEF3E7}

\definecolor{jointpurple}{HTML}{8E6FD5}
\definecolor{jointpurplebg}{HTML}{F1EEF9}
\newcommand{\purplehl}[1]{%
  \sethlcolor{jointpurplebg}%
  \textcolor{jointpurple}{\textbf{\textit{\hl{#1}}}}%
}

\newcommand{\newpara}[1]{\vspace{5pt}\noindent\textbf{#1}}
\usepackage{cuted}
\usepackage{caption}

\newcommand{\ours}{JointREPA}
\newcommand{\savis}{UNSAIDBench}
\newcommand{\shareatt}{underspecified shared factors}

\definecolor{cvprblue}{rgb}{0.21,0.49,0.74}
\usepackage[pagebackref,breaklinks,colorlinks,allcolors=cvprblue]{hyperref}

\def\confName{CVPR}
\def\confYear{2027}

\title{Beyond Co-Generation: \\ Rethinking Jointness in Joint Audio–Video Generation}

\author{
Jihoo Jung$^{1}$ \quad
Ji-Hoon Kim$^{2}$ \quad
Suyeon Lee$^{1}$ \quad
Gia Thinh Pham$^{1}$ \quad
Joon Son Chung$^{1}$\\
$^{1}$KAIST \qquad $^{2}$Chung-Ang University
}
\begin{document}
\maketitle

\begin{abstract}
Our aim is to advance the understanding of genuine audio-visual coordination in joint text-to-audio-video (T2AV) generation, with particular emphasis on semantic consistency. Evaluating audio-visual consistency in T2AV generation is nontrivial, as detailed text prompts obscure whether such consistency stems from genuine cross-modal coordination or merely from independent text following. To isolate the former, we introduce intrinsic jointness: the ability of audio and video to produce the same realization for details left underspecified by the prompt. Using semantically underspecified prompts (e.g., ``musical instrument playing'' rather than ``electric guitar playing''), we investigate four recent models and reveal their limited intrinsic jointness. To address this limitation, we propose JointREPA, a simple representation alignment-based post-training method to facilitate cross-modal coordination. Since existing benchmarks use semantically detailed prompts, we introduce UNSAIDBench, a benchmark designed to evaluate intrinsic jointness under semantic underspecification. Experiments across two distinct models and three benchmarks, including UNSAIDBench, demonstrate that JointREPA consistently improves semantic consistency.
\end{abstract}     

\section{Introduction}
\label{sec:intro}

Recent advances in text-to-image (T2I) and text-to-video (T2V) generation have enabled high-quality visual synthesis from natural language descriptions~\cite{xie2025sana,han2025infinity,ye2026ultraflux,chefer2025videojam,huang2025self,cui2026selfforcing}. Building on this progress, recent works have explored joint text-to-audio-video (T2AV) generation, in which both audio and video are synthesized from a shared text prompt~\cite{seedance2026seedance,sora2,veo31,wang2025universe,liu2026javisdit,liu2026javisdit++,low2025ovi,seedance2025seedance,wang2026apollo,hacohen2026ltx,kandinsky2026kandinsky}. Unlike single-modality generation tasks such as T2I and T2V, where faithfully following the text prompt is the main focus, T2AV additionally requires consistency between the two generated modalities. This cross-modal consistency encompasses two key aspects: \emph{semantic consistency}, which requires audio and video to depict the same events, and \emph{temporal consistency}, which requires corresponding events to occur at aligned times.

\begin{figure}[t]
    \centering
    \includegraphics[width=\linewidth]{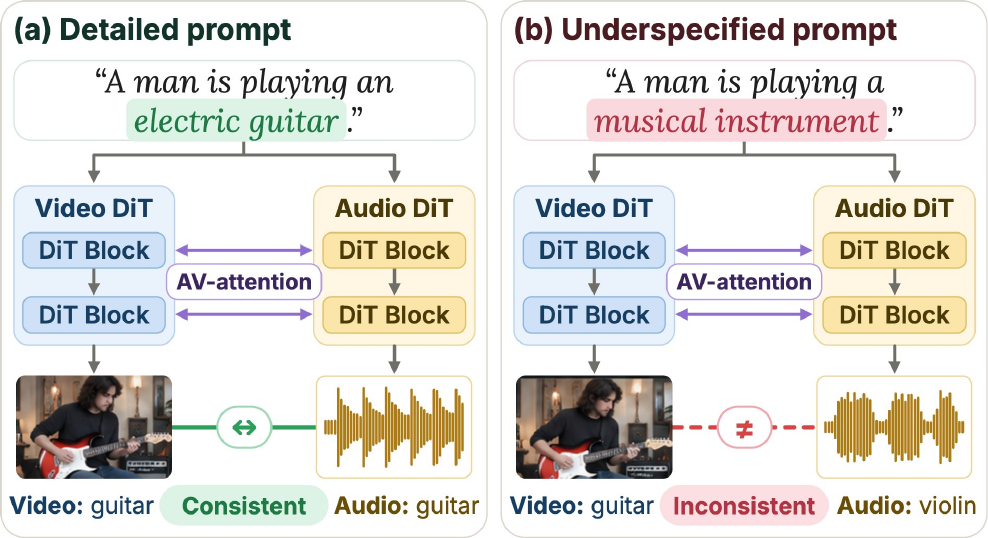}
    \vspace{-6mm}
        \caption{\textbf{Intrinsic jointness.} Under detailed prompts, output consistency can arise from each modality independently following the text prompt alone, without audio--visual interaction. In contrast, underspecified prompts enable disentangling whether this consistency stems from mere prompt following or genuine audio--visual coordination. We define intrinsic jointness as the ability to maintain consistency under such underspecified prompts.}
    \vspace{-6mm}
    \label{fig:main}
\end{figure}

While recent T2AV models appear to exhibit such consistency, this does not necessarily imply that the two modalities are genuinely coordinated. When the prompt specifies every detail, such consistency can arise simply from each modality following the prompt independently, even without cross-modal coordination. To evaluate cross-modal coordination beyond text following, we therefore examine \textbf{intrinsic jointness}: the ability of audio and video to converge on the same realization for details that the prompt leaves open to multiple possibilities. For example, an underspecified prompt such as ``A man is playing a \textit{musical instrument}'' (\cref{fig:main}(b)), which admits multiple valid realizations (\emph{e.g.}, guitar or violin), requires the two modalities to actively coordinate their choice to reliably produce consistent outputs. Intrinsic jointness matters in practice, as users rarely specify every detail in their prompts~\cite{cheng2025vpo,hahn2025proactive}. It is also important for physically coherent generation, where unstated object properties (\emph{e.g.}, size or material) should jointly shape both sound and visuals~\cite{hyun2026pavas}.

Existing benchmarks do not equally evaluate the semantic and temporal dimensions of intrinsic jointness. Since natural text prompts rarely state temporal properties such as exact timestamps~\cite{wu2025mint,xie2024picoaudio}, temporal coordination beyond text following has been widely examined. In contrast, semantic information is often specified in detail within standard benchmarks~\cite{liu2026javisdit,hua2025vabench,cao2025t2av}, leaving the semantic dimension largely underexplored. In this work, we focus specifically on this overlooked dimension of intrinsic jointness, hereafter referred to as \textit{semantic jointness}.


We investigate semantic jointness in four recent open-source T2AV models: LTX-2.3, LTX-2~\cite{hacohen2026ltx}, JavisDiT++~\cite{liu2026javisdit++}, and JavisDiT~\cite{liu2026javisdit}. We construct small diagnosis prompts across three levels of semantic specificity by progressively abstracting event identities (\emph{e.g.}, electric guitar $\rightarrow$ string instrument $\rightarrow$ musical instrument). Across all four models, semantic consistency between the generated audio and video drops substantially as prompts become less specific, revealing limited semantic jointness. Probing intermediate features shows that this inconsistency is not confined to the final output but persists throughout internal representations. By analyzing audio--visual attention, which enables cross-modal interaction, we further reveal that this pathway is insufficiently utilized to facilitate semantic coordination.

Motivated by these findings, we propose \textbf{\ours{}}, a simple representation alignment~\cite{yu2025representation} based post-training method for T2AV models. \ours{} aligns each modality's intermediate features with pretrained representations from both its own modality and the other modality, facilitating cross-modal coordination. We also introduce \textbf{\savis{}} (\textbf{UN}derspecified \textbf{S}emantics in \textbf{A}udio-v\textbf{ID}eo generation), a benchmark designed to evaluate semantic jointness. Unlike existing benchmarks that often explicitly specify semantic details, \savis{} features semantically underspecified prompts that admit multiple valid event realizations. Evaluated on \savis{} as well as the standard JavisBench~\cite{liu2026javisdit} and Verse-Bench~\cite{wang2025universe}, \ours{} yields consistent gains in audio--visual alignment across two distinct models.

Our contributions are threefold:
\begin{itemize}
    \item To assess genuine audio--visual coordination in T2AV models, we introduce intrinsic jointness, defined as the ability of audio and video to converge on the same realization for details that remain underspecified in the text prompt.
    \item Using semantically underspecified prompts, we systematically analyze four recent T2AV models, uncovering substantial limitations in semantic jointness.
    \item We propose \ours{}, a lightweight post-training method to promote audio--visual coordination, improving semantic consistency across two models and three benchmarks.
\end{itemize}


\section{Related Works}
\label{sec:related_work}

\subsection{Joint Text-to-audio-video Generation (T2AV)} 
T2AV generation aims to synthesize audio and video simultaneously from a text prompt~\cite{seedance2026seedance,sora2,veo31,wang2025universe,liu2026javisdit,liu2026javisdit++,low2025ovi,seedance2025seedance,wang2026apollo,hacohen2026ltx,kandinsky2026kandinsky}. Dual-tower models (e.g., LTX-2, LTX-2.3~\cite{hacohen2026ltx}, Ovi~\cite{low2025ovi}, and JavisDiT~\cite{liu2026javisdit}) employ separate Diffusion Transformer (DiT)~\cite{peebles2023scalable} towers for audio and video. Each DiT block comprises unimodal self-attention, text cross-attention, and audio--visual cross-attention. Alternatively, single-tower models such as JavisDiT++~\cite{liu2026javisdit++} flatten and concatenate audio and video tokens, feeding them into a shared DiT tower. Its DiT blocks consist of joint audio--visual self-attention, text cross-attention, and modality-specific FFNs. Most models adopt flow matching~\cite{lipman2023flow} as their primary training objective. While single-modality generation such as text-to-image (T2I) or text-to-video (T2V) primarily focuses on text alignment, T2AV generation additionally requires cross-modal consistency, both semantically and temporally. Although current models appear to achieve such consistency, we investigate whether this alignment truly stems from genuine cross-modal coordination.
 \begin{figure*}[t]
  \centering
  \includegraphics[width=0.8\textwidth]{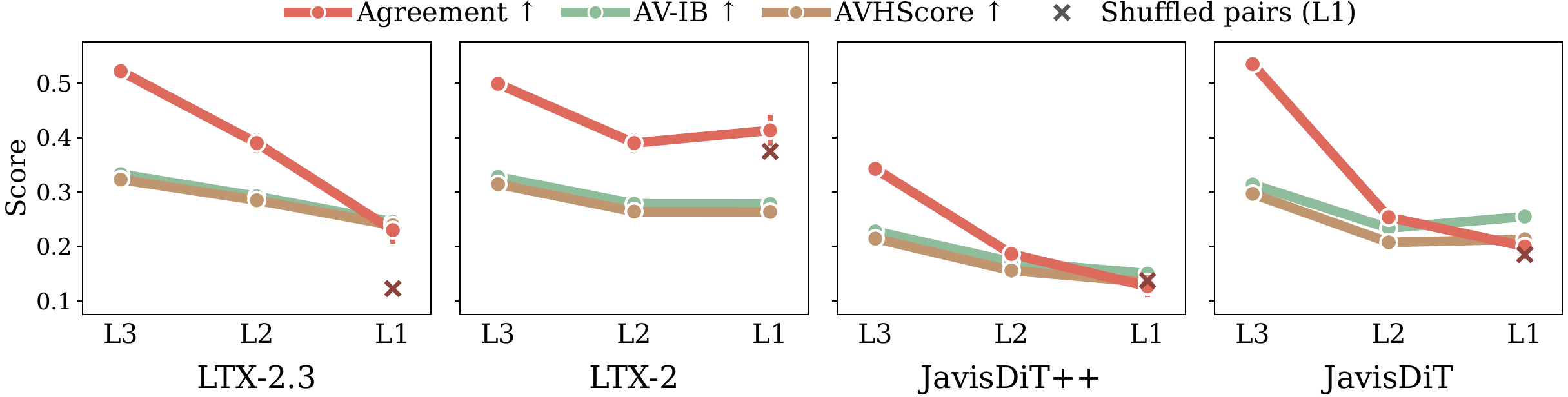}
  \vspace{-3mm}
\caption{\textbf{Q1. Semantic jointness under varying prompt specificity.}
We evaluate semantic consistency using three metrics (Agreement, AV-IB, and AVHScore) across different levels of prompt specificity.
All metrics generally decrease as prompt specificity decreases (L3 $\to$ L2 $\to$ L1). At L1, Agreement scores for most models are close to the shuffled-pair baseline (marked by $\times$), obtained by randomly pairing audio and video generated with different seeds under the same prompt. This suggests that current T2AV models exhibit limited semantic jointness.}
  \label{fig:prompt_specificity}
  \vspace{-5mm}
\end{figure*}
\subsection{Prompt Specificity in Generative Models}
Prompt specificity has been widely studied in T2I and T2V generation from two main perspectives. One line of work investigates the impact of prompt specificity on generation performance. Zhang et al.~\cite{zhang2026the} analyze how prompt complexity influences quality, diversity, and consistency in T2I generation. Liu et al.~\cite{liu2023fetv} categorizes prompts along multiple dimensions, including complexity, and evaluates T2V model performance across these dimensions. Another line of work views prompt underspecification as an issue to be mitigated. These approaches seek to resolve ambiguities left by the prompts by soliciting user clarifications~\cite{mehrabi-etal-2023-resolving,hahn2025proactive} or developing models for automatic prompt optimization~\cite{cheng2025vpo}. In T2AV generation, such underspecification poses a different challenge: when multiple valid realizations exist, audio and video are required to converge on the same one. In this work, we use underspecified prompts as a probe to investigate whether models genuinely coordinate across modalities.

\subsection{Representation Alignment for Generation} Aligning internal representations with pretrained encoders has recently been widely explored in generative models. REPA~\cite{yu2025representation} aligns hidden states in denoising networks with clean image representations from pretrained visual encoders during T2I model training. This encourages the model to capture useful semantic information, improving both training efficiency and generation quality. Recent studies have further adapted this paradigm to other generative domains: VideoREPA~\cite{zhang2025videorepa} extends it to T2V generation and proposes Token Relation Distillation to improve physical plausibility; AG-REPA~\cite{zhang2026agrepa} applies representation alignment to text-to-audio generation, introducing a causal layer selection strategy to enhance audio quality; and TARO~\cite{Ton_2025_ICCV} and MGAudio~\cite{zhang2025modelguided} adapt it to video-to-audio generation to improve cross-modal coherence and audio realism. Our work extends this paradigm to T2AV generation, addressing the semantic jointness bottleneck by aligning features both within and between the two modalities.

\section{Diagnosing Semantic Jointness}
\label{sec:rethinking_jointness}
In this section, we first formalize intrinsic jointness in Sec.~\ref{sec:problem_formulation}, outline our analysis setup in Sec.~\ref{sec:setup}, and present our findings in Sec.~\ref{sec:analysis}.

\subsection{Defining Intrinsic Jointness}
\label{sec:problem_formulation}

Given a text prompt $c$, let $p_\theta(a,v\mid c)$
denote the distribution of generated audio--video pairs.
Since prompts rarely specify every detail, sampling
from $p_\theta$ necessitates realizing the details left unspecified by $c$. Some such details are modality-specific (e.g., clothing color), while others must be resolved consistently across audio and video. We formalize the latter as a set of \emph{\shareatt{}} $\mathcal{Z}(c)=\{Z_1,\ldots,Z_K\}$.
For example, in \cref{fig:main}(b), these factors
include instrument type ($Z_1$), playing timing
($Z_2$) and others.
Each $Z_k$ is drawn from a set of possible realizations $\mathcal{R}_k(c)$ (\emph{e.g.}, $\mathcal{R}_1(c)=\{\text{guitar},\text{violin},\dots\}$), forming the joint realization space $\mathcal{R}(c) = \mathcal{R}_1(c) \times \dots \times \mathcal{R}_K(c)$. 

Let $\mathbf{z}^a = (z_1^a, \dots, z_K^a)$ and $\mathbf{z}^v = (z_1^v, \dots, z_K^v) \in \mathcal{R}(c)$ denote their realizations in audio and video, respectively. We view joint generation $p_\theta(a,v\mid c)$ as realizing the \shareatt{} as $\mathbf{z}^a$ and $\mathbf{z}^v$ in each modality.
Audio and video are generated conditioned on $c$ and their respective realizations:
\begin{equation}
\begin{aligned}
p_\theta(a,v\mid c)
&= \sum_{\mathbf{z}^a,\mathbf{z}^v}
\pi_\theta(\mathbf{z}^a,\mathbf{z}^v\mid c)\\
&\quad\cdot
p_\theta(a\mid c,\mathbf{z}^a)
p_\theta(v\mid c,\mathbf{z}^v),
\end{aligned}
\label{eq:joint_realization}
\end{equation}
where $\pi_\theta(\mathbf{z}^a,\mathbf{z}^v\mid c)$
is the joint distribution over the \shareatt{} realized by the two modalities.

Ideally, a T2AV model resolves \shareatt{} consistently across audio and video, yielding the same realization in both modalities, i.e., $\mathbf{z}^a=\mathbf{z}^v$.
We define \textbf{intrinsic jointness} under prompt $c$ as the probability that audio and video agree on the \shareatt{}:
\begin{equation}
J_\theta(c)
=\Pr_\theta\!\left(\mathbf{z}^a=\mathbf{z}^v\mid c\right)
=\sum_{\mathbf{z}\in\mathcal{R}(c)}\pi_\theta(\mathbf{z},\mathbf{z}\mid c).
\label{eq:jointness}
\end{equation}

\begin{figure*}[t]
    \centering
    \includegraphics[width=0.99\textwidth]
    {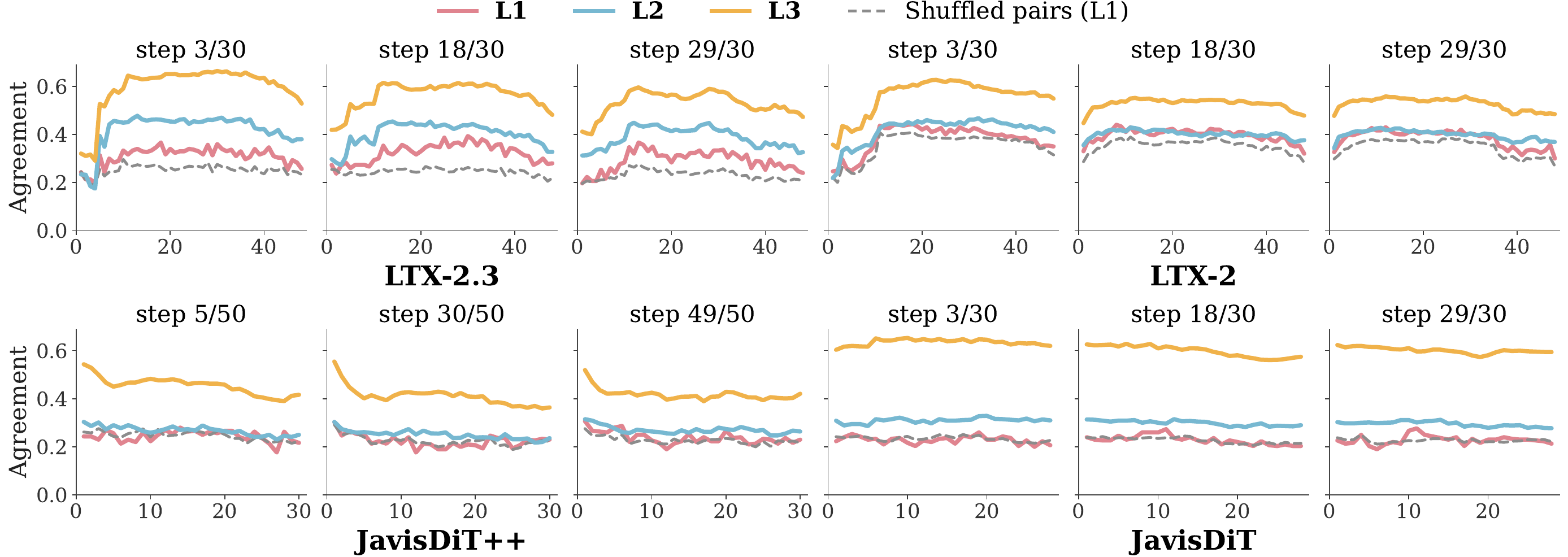}
    \vspace{-2mm}
    \caption{\textbf{Q2. Semantic jointness in internal representations.}
We evaluate probe agreement across Transformer blocks (horizontal axis) at different generation steps and levels of prompt specificity.
Probe agreement generally decreases as prompt specificity decreases (L3 $\to$ L2 $\to$ L1). At L1, probe agreement is close to the shuffled-pair baseline (dashed line). This suggests that limited semantic jointness is also reflected in the models' internal representations.}
    \label{fig:representation_probe}
    \vspace{-6mm}
\end{figure*} 

We focus on the semantic aspect of intrinsic jointness. That is, \shareatt{} in prompt $c$ concern semantic attributes, such as source or event identity, rather than temporal properties, such as event timing. Because natural text prompts often omit exact timestamps~\cite{wu2025mint,xie2024picoaudio}, temporal coordination has been widely examined. In contrast, semantic information is typically specified in detail~\cite{liu2026javisdit,hua2025vabench,cao2025t2av}, leaving this semantic dimension---hereafter referred to as \emph{semantic jointness}---largely underexplored.

\subsection{Analysis Setup}
\label{sec:setup}
\newpara{Data.} To evaluate semantic jointness, we construct prompts $c$ with exactly one underspecified shared factor ($\mathcal{Z}(c) = \{Z_1\}$). For this factor, we focus on events clearly distinguishable in both audio and video~\cite{zhou2025towards}, choosing three categories: \emph{musical instrument playing}, \emph{human speech}, and \emph{animal vocalization} (\cref{tab:prompt_levels}). Guided by the AudioSet ontology~\cite{7952261}, we organize each event into three levels of semantic specificity (L1--L3) such that the size of the valid realization set $|\mathcal{R}(c)|$ progressively shrinks: from a broad category (L1, \emph{e.g.}, \emph{musical instrument}) to a coarse class (L2, \emph{e.g.}, \emph{string instrument}) and a fine-grained event (L3, \emph{e.g.}, \emph{guitar}). We construct the test prompts by specifying a single underspecified shared factor at each level while keeping the surrounding text fixed (\emph{e.g.}, \emph{``A person is playing a \_\_\_.''}). 


\begin{table}[t]
\centering
\footnotesize
\setlength{\tabcolsep}{4pt}
\renewcommand{\arraystretch}{1.12}
\newcommand{\blank}{\rule[-0.3ex]{1.5em}{0.5pt}}
\newcommand{\sep}{\,{\color{gray}/}\,}
\newcommand{\tmpl}[1]{\rowcolor{gray!12}\multicolumn{3}{@{\hspace{3pt}}l}{\textit{``#1''}}\\}
\begin{tabular}{@{}>{\raggedright\arraybackslash}p{0.21\columnwidth}
                  >{\raggedright\arraybackslash}p{0.30\columnwidth}
                  >{\raggedright\arraybackslash}p{0.42\columnwidth}@{}}
\toprule
\textbf{L1} & \textbf{L2} & \textbf{L3} \\
\midrule
\tmpl{A/An \blank\ is vocalizing.}
\multirow{3}{=}{animal}
  & pet         & dog \sep cat \\
  & farm animal & cow \sep sheep \\
  & wild animal & lion \sep wolf \\
\cmidrule(l{0pt}r{0pt}){1-3}
\tmpl{A person is playing a \blank.}
\multirow{3}{=}{musical instrument}
  & string inst.     & guitar \sep violin \\
  & percussion inst. & drum \sep xylophone \\
  & wind inst.       & flute \sep trumpet \\
\cmidrule(l{0pt}r{0pt}){1-3}
\tmpl{A/An \blank\ is speaking.}
\multirow{2}{=}{person}
  & man & adult man \sep young boy \\
  & woman & adult woman \sep young girl \\
\bottomrule
\end{tabular}
\vspace{-3mm}
\caption{\textbf{Simple diagnostic set for semantic jointness analysis.} Prompts at L1, L2, and L3 share the same template but differ in the level of semantic specificity.}
\vspace{-5mm}
\label{tab:prompt_levels}
\end{table}

\newpara{Models.} For each prompt, we generate 100 samples using different random seeds with four recent open-source T2AV models: three dual-tower DiTs (LTX-2.3, LTX-2~\cite{hacohen2026ltx}, JavisDiT~\cite{liu2026javisdit}) and one single-tower DiT (JavisDiT++~\cite{liu2026javisdit++}). Dual-tower models couple the two modalities through audio--visual cross-attention, whereas the single-tower model applies joint self-attention over concatenated audio and video tokens. For simplicity, we refer to these attention between audio and video tokens in either architecture as \textit{AV-attention}. All models use flow matching~\cite{lipman2023flow} as their base objective.

\newpara{Metrics.} To evaluate semantic jointness, we operationalize $\mathbf{z}^a$ and $\mathbf{z}^v$ in \cref{eq:jointness} via classification over predefined realization space $\mathcal{R}(c)$. Specifically, we employ category- and modality-specific classifiers (\emph{e.g.}, SigLIP~\cite{zhai2023sigmoid} and LAION-CLAP~\cite{wu2023large} for the video and audio streams of musical instruments, respectively) to independently predict the realized identities $\hat{\mathbf{z}}^a, \hat{\mathbf{z}}^v$. We then measure identity \textit{agreement} over $N$ generated samples as $\frac{1}{N}\sum_{i=1}^{N} \mathbb{I}(\hat{\mathbf{z}}_i^a = \hat{\mathbf{z}}_i^v)$. We also report two standard audio--visual semantic consistency metrics, AV-IB~\cite{girdhar2023imagebind} and AVHScore~\cite{TAVGBench}.

\subsection{Findings}
\label{sec:analysis}
\newpara{Q1. Do existing T2AV models exhibit semantic jointness?} \cref{fig:prompt_specificity} plots Agreement, AV-IB, and AVHScore across specificity levels (L1--L3) for four models. We make two observations. First, all metrics generally degrade as prompts become less specific (L3 $\to$ L2 $\to$ L1). Second, for most models, L1 Agreement closely approaches a shuffled baseline that randomly pairs audio and video generated from different seeds within the same text prompt. Together, these results indicate that the two modalities resolve unspecified identities nearly independently, struggling to achieve true semantic jointness.

\newpara{Q2. Does limited semantic jointness persist in internal representations?} We examine whether this disagreement arises only at the final output or persists within internal representations. To this end, we measure internal agreement by probing intermediate features across Transformer blocks at various denoising steps. Specifically, at each block and step, we separately mean-pool audio and video tokens and train modality-specific linear probes to predict each modality's output identity label. We then measure the agreement between the two modalities' probe predictions (\cref{fig:representation_probe}). Probe agreement generally decreases as prompts become less specific, with the L1 curves remaining close to the shuffled baseline. These findings indicate that limited semantic jointness persists in internal representations.

\begin{figure*}[t]
    \centering
    
    \includegraphics[width=\textwidth]{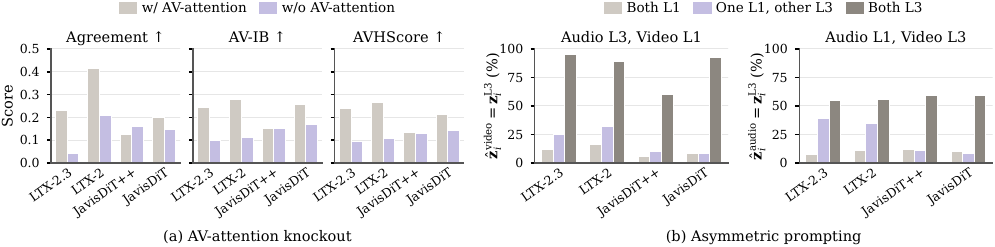}
    \vspace{-7mm}
   \caption{\textbf{Q3. Analysis of the role of AV-attention in semantic coordination.}
(a) Semantic consistency with and without AV-attention. Blocking AV-attention causes substantial drops across all metrics in the LTX family, while inducing only minor changes in the JavisDiT family.
(b) Semantic identity following under asymmetric prompting (L3 for one modality, L1 for the other). For both model families, the L1-conditioned modality $m_{\mathrm{L1}}$ follows the target identity specified $\mathbf{z}^{\mathrm{L3}}$ substantially less reliably than under symmetric L3 prompting (both-L3). Together, these findings suggest that AV-attention is not effectively utilized for semantic coordination.}
    \vspace{-5mm}
    \label{fig:avcut_follow}
\end{figure*}

\newpara{Q3. How effective is AV-attention for semantic coordination?}
Since cross-modal information exchange in T2AV models occurs through AV-attention, we conduct two experiments to examine its role in semantic coordination.

\noindent{\bf AV-attention knockout.} First, we examine whether T2AV models actively utilize AV-attention during generation. Following attention knockout~\cite{geva2023dissecting}, we block AV-attention at inference time and measure how much semantic consistency degrades. As shown in \cref{fig:avcut_follow}(a), blocking AV-attention causes substantial drops across all metrics for the LTX family. In contrast, JavisDiT exhibits only a slight decline, and JavisDiT++ remains nearly unchanged. This indicates that the audio-visual pathway is active in the LTX family, but largely underutilized in the JavisDiT family.

\noindent{\bf Asymmetric prompting.}
Next, we test how effectively this pathway carries semantic information across modalities. Specifically, we condition one modality $m_{\mathrm{L3}}$ on an L3 prompt specifying a target identity $\mathbf{z}^{\mathrm{L3}}$ and the other modality $m_{\mathrm{L1}}$ on a broader L1 prompt, and measure the \textit{follow rate}---the fraction of samples in which the predicted identity of L1-conditioned modality $\hat{\mathbf{z}}_i^{m_{\mathrm{L1}}}$ matches $\mathbf{z}_i^{\mathrm{L3}}$: $\frac{1}{N}\sum_{i=1}^{N} \mathbb{I}(\hat{\mathbf{z}}_i^{m_{\mathrm{L1}}} = \mathbf{z}_{i}^{\mathrm{L3}})$. In \cref{fig:avcut_follow}(b), the left plot feeds L3 prompt to audio and L1 to video to measure $\frac{1}{N}\sum_{i=1}^{N} \mathbb{I}(\hat{\mathbf{z}}_i^{\mathrm{video}} = \mathbf{z}_{i}^{\mathrm{L3}})$, and vice versa on the right. For comparison, we also reports symmetric prompting baselines, where both modalities are conditioned on the same prompt level: either L1 (both-L1) or L3 (both-L3). As in the asymmetric setting, we evaluate the fraction of samples where the predicted identity $\hat{\mathbf{z}}_i^{m}$ of each modality matches the target $\mathbf{z}_i^{\mathrm{L3}}$, given by $\frac{1}{N}\sum_{i=1}^{N} \mathbb{I}(\hat{\mathbf{z}}_i^{m} = \mathbf{z}_i^{\mathrm{L3}})$
($m=\mathrm{video}$ for the left panel and $m=\mathrm{audio}$ for the right panel).

The LTX family exhibits a higher follow rate than the both-L1 baseline, indicating that semantic information can be transferred across modalities through AV-attention. However, the follow rate remains substantially lower than in the both-L3 setting, suggesting that AV-attention is not fully effective in facilitating semantic coordination. The JavisDiT family exhibits negligible gains over the both-L1 baseline in either direction, with follow rates remaining substantially below those in the both-L3 setting.

Overall, despite varying degrees of effectiveness across models, AV-attention remains limited in its ability to transfer semantic information across modalities in all four models.

\section{Improving Semantic Jointness}
\label{sec:method}

\begin{figure}[t]
    \centering
    \includegraphics[width=\columnwidth]{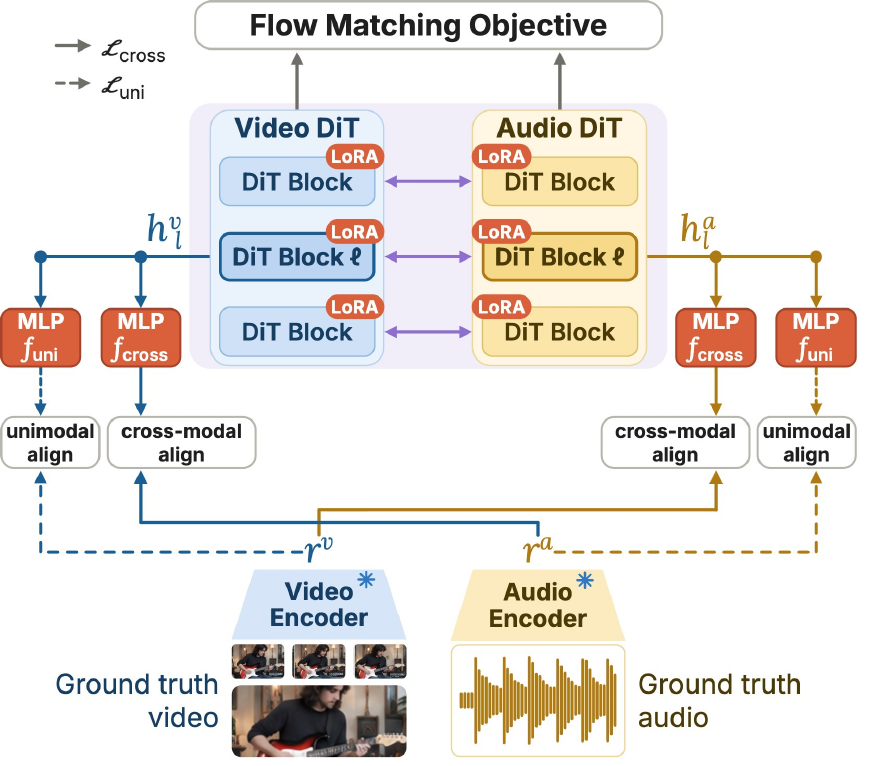}
    \vspace{-6mm}
\caption{\textbf{Overview of \ours{}.}
\ours{} aligns each modality's features with pretrained encoder representations from both its own modality and the other modality.}
    \vspace{-6mm}
    \label{fig:method}
\end{figure}

Motivated by the findings in \cref{sec:analysis}, we propose \textbf{\ours{}}, which extends REPA~\cite{yu2025representation} to T2AV generation. \ours{} aligns each modality's representation not only with target features from its own pretrained encoder, but also with those from the other modality. The former builds semantically rich representations that provide clearer signals for AV-attention to leverage, while the latter actively promotes cross-modal semantic exchange.

\begin{table*}[t]
\centering
\caption{\textbf{Main results on \savis{}, JavisBench-mini, and VerseBench.} We apply \ours{} to JavisDiT++ and LTX-2.3 and evaluate AV-Consistency, AV-Synchrony, Text-Consistency, and AV-Quality on three benchmarks.
\ours{} consistently improves AV-Consistency across all benchmarks and backbone models.}
\vspace{-2mm}
\label{tab:main_results}
\resizebox{\textwidth}{!}{%
\begin{tabular}{l l cc cc cccc cc}
\toprule
 & & \multicolumn{2}{c}{AV-Consistency}
 & \multicolumn{2}{c}{AV-Synchrony}
 & \multicolumn{4}{c}{Text-Consistency}
 & \multicolumn{2}{c}{AV-Quality} \\
\cmidrule(lr){3-4}
\cmidrule(lr){5-6}
\cmidrule(lr){7-10}
\cmidrule(lr){11-12}
Model & Size
& AV-IB $\uparrow$
& AVHScore $\uparrow$
& JavisScore $\uparrow$
& DeSync $\downarrow$
& TV-IB $\uparrow$
& TA-IB $\uparrow$
& CLIP $\uparrow$
& CLAP $\uparrow$
& VQ $\uparrow$
& AQ $\uparrow$ \\
\midrule

\multicolumn{12}{c}{\textbf{\savis{}}} \\
\midrule

JavisDiT++~\citep{liu2026javisdit++}
& 2.1B
& 0.176 & 0.163
& 0.133 & 0.815
& \textbf{0.180} & 0.138 & \textbf{0.313} & 0.343
& 1.455 & 4.595 \\
\rowcolor{repa_blue}
\quad + \textbf{\ours{}}
& 2.1B
& \textbf{0.207} & \textbf{0.196}
& \textbf{0.164} & \textbf{0.804}
& 0.174 & \textbf{0.163} & 0.310 & \textbf{0.369}
& \textbf{1.655} & \textbf{4.628} \\

\midrule

LTX-2.3~\citep{hacohen2026ltx}
& 22B
& 0.244 & 0.236
& 0.209 & 0.468
& \textbf{0.173} & 0.101 & 0.298 & 0.287
& \textbf{2.354} & \textbf{5.050} \\
\rowcolor{repa_blue}
\quad + \textbf{\ours{}}
& 22B
& \textbf{0.302} & \textbf{0.288}
& \textbf{0.261} & \textbf{0.440}
& 0.169 & \textbf{0.124} & 0.298 & \textbf{0.327}
& 2.256 & 4.852 \\

\midrule

\multicolumn{12}{c}{\textbf{JavisBench-mini~\cite{liu2026javisdit}}} \\
\midrule

JavisDiT++~\citep{liu2026javisdit++}
& 2.1B
& 0.200 & 0.182
& 0.157 & 0.784
& \textbf{0.283} & 0.170 & 0.316 & 0.429
& 1.538 & \textbf{4.987} \\

\rowcolor{repa_blue}
\quad + \textbf{\ours{}}
& 2.1B
& \textbf{0.221} & \textbf{0.204}
& \textbf{0.179} & \textbf{0.782}
& 0.278 & \textbf{0.187} & 0.316 & \textbf{0.447}
& \textbf{1.715} & 4.917 \\

\midrule

LTX-2.3~\citep{hacohen2026ltx}
& 22B
& 0.249 & 0.238
& 0.214 & \textbf{0.555}
& \textbf{0.280} & 0.171 & \textbf{0.306} & 0.412
& \textbf{2.458} & \textbf{5.348} \\

\rowcolor{repa_blue}
\quad + \textbf{\ours{}}
& 22B
& \textbf{0.285} & \textbf{0.271}
& \textbf{0.247} & 0.575
& 0.277 & \textbf{0.186} & 0.304 & \textbf{0.433}
& 2.339 & 5.157 \\

\midrule

\multicolumn{12}{c}{\textbf{VerseBench~\cite{wang2025universe}}} \\
\midrule

JavisDiT++~\citep{liu2026javisdit++}
& 2.1B
& 0.213 & 0.201
& 0.172 & \textbf{0.905}
& \textbf{0.182} & 0.189 & \textbf{0.325} & 0.479
& 1.580 & \textbf{4.280} \\

\rowcolor{repa_blue}
\quad + \textbf{\ours{}}
& 2.1B
& \textbf{0.237} & \textbf{0.223}
& \textbf{0.195} & 0.958
& 0.175 & \textbf{0.216} & 0.323 & \textbf{0.489}
& \textbf{1.850} & 4.260 \\

\midrule

LTX-2.3~\citep{hacohen2026ltx}
& 22B
& 0.332 & 0.317
& 0.287 & 0.284
& \textbf{0.167} & 0.191 & 0.317 & 0.447
& \textbf{2.760} & \textbf{5.004} \\

\rowcolor{repa_blue}
\quad + \textbf{\ours{}}
& 22B
& \textbf{0.369} & \textbf{0.351}
& \textbf{0.322} & \textbf{0.241}
& 0.165 & \textbf{0.211} & 0.317 & \textbf{0.472}
& 2.677 & 4.809 \\

\bottomrule
\end{tabular}%
}
\vspace{-1mm}
\end{table*}

\begin{figure*}[t]
\vspace{1mm}
    \centering
    \includegraphics[width=\textwidth]{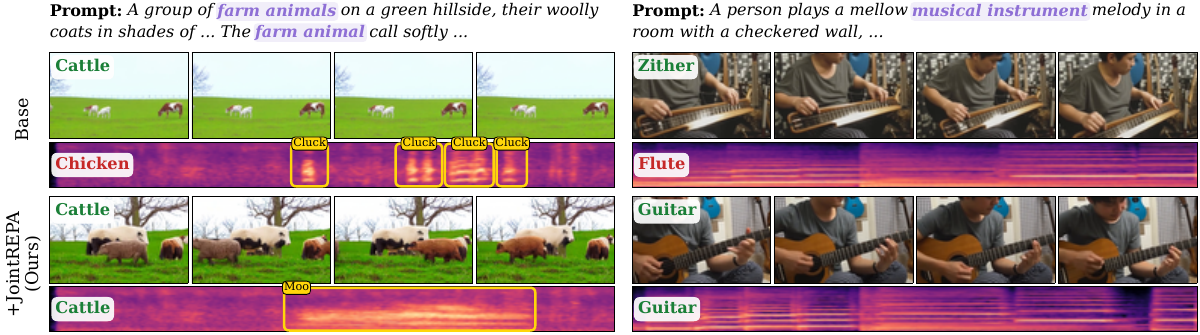}
    \vspace{-6mm}
    \caption{\textbf{Qualitative results on \savis{}.}
Generated video frames and audio spectrograms from the base model (top) and its \ours{} variant (bottom). Given an \purplehl{underspecified shared factor} per prompt, the base model realizes different semantic identities in audio and video, whereas \ours{} converges on the same identity.}
\vspace{-4mm}
    \label{fig:qual}
\end{figure*}

As illustrated in \cref{fig:method}, let $h^a_{\ell}$ and $h^v_{\ell}$ denote the pooled audio and video hidden states at layer $\ell$, and $r^a$ and $r^v$ the target features from frozen pretrained encoders (CLAP~\cite{10095889} for audio and V-JEPA~\cite{assran2025v} for video). Each modality uses two MLP projectors, $f_{\mathrm{uni}}$ and $f_{\mathrm{cross}}$, to align its hidden states with the unimodal target and cross-modal counterpart, respectively. The unimodal ($\mathcal{L}_{\mathrm{uni}}$) and cross-modal ($\mathcal{L}_{\mathrm{cross}}$) alignment losses are formulated as:
\begin{equation}
\begin{aligned}
\mathcal{L}_{\mathrm{uni}}
= -\mathbb{E}\Big[
&\operatorname{sim}\big(
f_{\mathrm{uni}}^a(h^a_\ell), r^a\big)\\
&+\operatorname{sim}\big(
f_{\mathrm{uni}}^v(h^v_\ell), r^v\big)
\Big], \\[1ex]
\mathcal{L}_{\mathrm{cross}}
= -\mathbb{E}\Big[
&\operatorname{sim}\big(
f_{\mathrm{cross}}^a(h^a_\ell), r^v\big)\\
&+\operatorname{sim}\big(
f_{\mathrm{cross}}^v(h^v_\ell), r^a\big)
\Big],
\end{aligned}
\label{eq:repa}
\end{equation}
where $\mathrm{sim}(\cdot, \cdot)$ denotes a similarity function.

The overall training objective is
\begin{equation}
    \mathcal{L} = \mathcal{L}_{\mathrm{FM}} + \lambda_{\mathrm{uni}}\mathcal{L}_{\mathrm{uni}} + \lambda_{\mathrm{cross}}\mathcal{L}_{\mathrm{cross}},
    \label{eq:total_loss}
\end{equation}
where $\mathcal{L}_{\mathrm{FM}}$ is the flow-matching loss, and $\lambda_{\mathrm{uni}}$ and $\lambda_{\mathrm{cross}}$ weight the alignment terms.

\section{Experiments}

\subsection{Experimental Setup}

\newpara{Datasets.} For training, we use a balanced subset of VGGSound~\cite{chen2020vggsound} across 52 classes. We exclude the categories used in \cref{sec:setup} to ensure a fair evaluation when revisiting the analysis in \cref{sec:add}. Existing benchmarks provide detailed semantic information, making it difficult to evaluate semantic jointness~\cite{liu2026javisdit,hua2025vabench,cao2025t2av}. To this end, we introduce \textbf{\savis{}} (\textbf{UN}derspecified \textbf{S}emantics in \textbf{A}udio-v\textbf{ID}eo generation). Moving beyond the simple diagnostic prompts in \cref{tab:prompt_levels}, we adapt the complex, real-world prompts of JavisBench~\cite{liu2026javisdit} to inject one underspecified shared factor per prompt. Specifically, following the AudioSet ontology~\cite{7952261}, we abstract fine-grained event identities into their parent categories while keeping the surrounding context intact, yielding 524 instances that remain natural in context.
To assess the generalizability of \ours{}, we additionally evaluate it on two standard benchmarks, JavisBench-mini and Verse-Bench~\cite{wang2025universe}. Since Verse-Bench uses both text and image conditioning, which JavisDiT++ does not support, we evaluate all models using only its text prompts.

\newpara{Evaluation metrics.} Following prior works~\cite{zhang2026omninft,xu2026av}, we evaluate four complementary dimensions:
(1)~\textit{AV-Consistency}, measuring
cross-modal semantic coherence using
AV-IB~\citep{girdhar2023imagebind} and AVHScore~\cite{TAVGBench};
(2)~\textit{AV-Synchronization}, evaluating
temporal alignment using
JavisScore~\citep{liu2026javisdit} and DeSync~\cite{iashin2024synchformer, cheng2025mmaudio}.
(3)~\textit{AV-Quality}, assessing the perceptual quality of each modality independently using
Visual Quality (VQ) from VideoAlign~\citep{liu2026improving} and
Audio Quality (AQ) from Audiobox-Aesthetics~\citep{tjandra2025meta};
(4)~\textit{Text-Consistency}, assessing
text-to-modality alignment using
ImageBind~\citep{girdhar2023imagebind},
CLIP~\citep{radford2021learning}, and CLAP~\cite{10095889}.

\newpara{Implementation details.} We select one representative model from each model family, JavisDiT++ and LTX-2.3, as our backbones. We apply \ours{} to both models using LoRA~\cite{hu2022lora} with a rank of 32.
We train for one epoch using AdamW with a learning rate of $10^{-4}$. We use effective batch sizes of 16 for JavisDiT++ and 4 for LTX-2.3. For the similarity function in \cref{eq:repa}, we employ token-relation distillation for unimodal video alignment following VideoREPA~\cite{zhang2025videorepa}, while using cosine similarity for the remaining terms. 
We apply \ours{} at the middle Transformer layer ($\ell=15$ out of 30 layers for JavisDiT++ and $\ell=24$ out of 48 layers for LTX-2.3), with $\lambda_{\mathrm{uni}}=\lambda_{\mathrm{cross}}=0.5$.

\subsection{Main Results}
\newpara{Quantitative results.}
\cref{tab:main_results} summarizes the quantitative results across \savis{}, JavisBench, and Verse-Bench.
Across both JavisDiT++ and LTX-2.3, \ours{} consistently improves AV-Consistency metrics over the respective base models. The gains are most pronounced on \savis{}, suggesting that \ours{} effectively guides both modalities to converge on the same identity among the multiple valid realizations. Notably, these improvements also extend to JavisBench and Verse-Bench, highlighting the general effectiveness of \ours{} in enhancing semantic consistency across diverse prompting conditions. Furthermore, \ours{} improves AV-Synchrony metrics in most settings, while yielding mixed results in Text-Consistency and AV-Quality. These results suggest that the benefits of \ours{} primarily manifest in cross-modal coordination.

\newpara{Qualitative results.}
\cref{fig:qual} presents qualitative results on \savis{}. In the animal vocalization example, the base model generates a video of cattle accompanied by chicken clucking sounds. Similarly, in the musical instrument example, the video depicts a zither being played, while the audio contains flute sounds. Although each modality independently satisfies the underspecified prompt, the generated pairs are semantically inconsistent. In contrast, \ours{} generates matching cattle vocalizations and guitar sounds for the corresponding visual events, yielding semantically coherent audio--video pairs. Additional samples are provided in the supplementary material.

\begin{table}[!htbp]
\centering
\small
\setlength{\tabcolsep}{3.5pt}

\caption{\textbf{Ablations of fine-tuning losses on JavisDiT++.} While the FM loss yields modest improvements and unimodal REPA provides larger gains, \ours{} achieves the best performance.}
\label{tab:ablation_results}
\vspace{-1mm}

\resizebox{\columnwidth}{!}{%
\begin{tabular}{lcccc}
\toprule
& \multicolumn{2}{c}{AV-Consistency}
& \multicolumn{2}{c}{AV-Synchrony} \\
\cmidrule(lr){2-3} \cmidrule(lr){4-5}
Model
& AV-IB$\uparrow$ & AVHScore$\uparrow$
& JavisScore$\uparrow$ & DeSync$\downarrow$ \\
\midrule

\multicolumn{5}{c}{\textbf{\savis{}}} \\
\midrule

JavisDiT++~\citep{liu2026javisdit++}
& 0.176 & 0.163 & 0.133 & \underline{0.815} \\

\quad + FM loss
& 0.181 & 0.171 & 0.141 & 0.826 \\

\quad + Unimodal REPA
& \underline{0.199} & \underline{0.188}
& \underline{0.158} & 0.832 \\

\rowcolor{repa_blue}
\quad + \textbf{\ours{}} (Ours)
& \textbf{0.207} & \textbf{0.196}
& \textbf{0.164} & \textbf{0.804} \\
\midrule

\quad + BEATs \ours{}
& 0.213 & 0.201 & 0.170 & 0.818 \\

\midrule
\multicolumn{5}{c}{\textbf{JavisBench-mini}} \\
\midrule

JavisDiT++~\citep{liu2026javisdit++}
& 0.200 & 0.182 & 0.157 & 0.784 \\

\quad + FM loss
& 0.204 & 0.187 & 0.160 & \textbf{0.780} \\

\quad + Unimodal REPA
& \underline{0.215} & \underline{0.199}
& \underline{0.174} & 0.787 \\

\rowcolor{repa_blue}
\quad + \textbf{\ours{}} (Ours)
& \textbf{0.221} & \textbf{0.204}
& \textbf{0.179} & \underline{0.782} \\
\midrule

\quad + BEATs \ours{}
& 0.221 & 0.204 & 0.179 & 0.772 \\

\bottomrule
\end{tabular}%
}

\vspace{3mm}

\begin{minipage}{\columnwidth}
    \centering
\includegraphics[width=\linewidth]{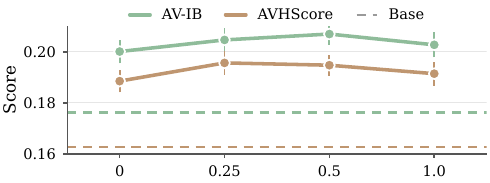}
        \vspace{-3mm}
        \caption{\textbf{Ablation on the cross-modal alignment weight \boldmath$\lambda_{\mathrm{cross}}$}. All non-zero weights consistently improve over $\lambda_{\mathrm{cross}}=0$, demonstrating robustness to $\lambda_{\mathrm{cross}}$.}
        \label{fig:lambda_cross}
        \vspace{-3mm}
    \end{minipage}
\vspace{3mm}

\begin{minipage}{\columnwidth}
    \centering
\includegraphics[width=\linewidth]{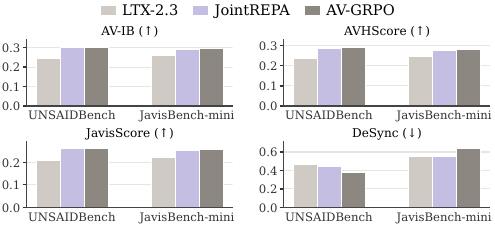}
        \vspace{-3mm}
        \caption{\textbf{Comparison of \ours{} and AV-GRPO~\cite{xu2026av} on LTX-2.3.}
        Despite relying on lightweight LoRA-based representation alignment rather than full-model reinforcement learning, \ours{} achieves comparable results with AV-GRPO.}
        \label{fig:av_grpo}
        \vspace{-3mm}
    \end{minipage}
\vspace{-3mm}
\end{table}
 
\begin{figure*}[t]
    \centering
    \vspace{1mm}
\includegraphics[width=\textwidth]{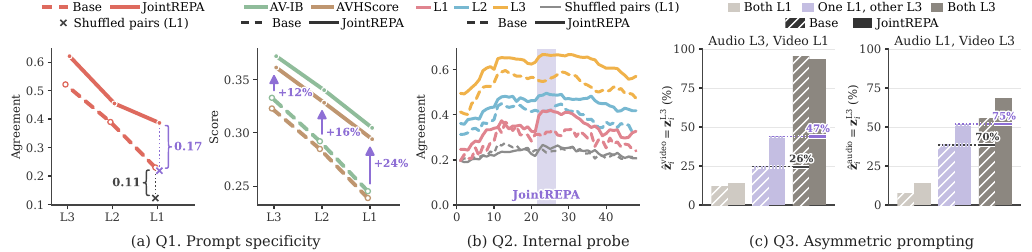}
    \vspace{-5mm}
    \caption{\textbf{Revisiting Q1, Q2 and Q3 after applying \ours{}.}
    (a) Q1. \ours{} improves all jointness metrics across specificity levels, peaking at L1—widening the Agreement margin over the shuffled baseline ($0.11 \rightarrow 0.17$) and achieving larger average gains in AV-IB and AVHScore at lower specificity ($\text{L3} < \text{L2} < \text{L1}$).
    (b) Q2. Intermediate representation probing reveals a pronounced increase in L1 Agreement around the middle layers where \ours{} is applied.
    (c) \ours{} increases the follow rate under asymmetric prompting (relative to Both-L3) from 26\% to 47\% and from 70\% to 75\%. }
    \label{fig:revisit}
    \vspace{-5mm}
\end{figure*}
\subsection{Additional Analysis} 
\label{sec:add}
\newpara{Ablations on fine-tuning loss.} In \cref{tab:ablation_results}, we conduct an ablation study of \ours{} on JavisDiT++ across \savis{} and JavisBench-mini. Fine-tuning with the flow-matching objective alone yields only marginal improvements. While applying unimodal alignment clearly improves AV-Consistency, combining it with cross-modal alignment in \ours{} achieves the best overall performance. To further examine the impact of the target representation, we substitute the default CLAP audio encoder with BEATs~\cite{pmlr-v202-chen23ag}. The resulting performance remains comparable, indicating that \ours{} is agnostic to different types of audio encoders. 

\newpara{Ablations on hyperparameter selection.} To examine the effect of cross-modal alignment, we train JavisDiT++ with varying $\lambda_{\mathrm{cross}}$. We report AV-Consistency on \savis{} in \cref{fig:lambda_cross}. Specifically, we vary $\lambda_{\mathrm{cross}}\in\{0, 0.25, 0.5, 1.0\}$ while fixing $\lambda_{\mathrm{uni}}=0.5$. Removing the cross-modal alignment term ($\lambda_{\mathrm{cross}}=0$) yields the weakest performance among the tested settings. In contrast, all non-zero weights yield consistently higher performance, with $\lambda_{\mathrm{cross}}=0.5$ achieving the best average performance across the two metrics. This indicates that \ours{} is robust to the selection of the cross-modal alignment weight $\lambda_{\mathrm{cross}}$.

\newpara{Comparison with other post-training methods.}
To assess the effectiveness of \ours{} relative to existing post-training approaches, we compare it with AV-GRPO~\cite{xu2026av} on both \savis{} and JavisBench-mini. AV-GRPO applies Group Relative Policy Optimization~\cite{guo2025deepseek} using modality quality, prompt alignment, and audio--video synchronization scores as rewards. We use its publicly released, fully fine-tuned LTX-2.3 checkpoint. As shown in \cref{fig:av_grpo}, even without explicit reward definitions or full-model fine-tuning, \ours{} achieves comparable AV-Consistency and AV-Synchrony on both benchmarks.

\newpara{Revisiting Q1, Q2, and Q3.}
We repeat the analyses of Q1, Q2, and Q3 from \cref{sec:analysis} after applying \ours{} to LTX-2.3. For Q1 (\cref{fig:revisit}(a)), \ours{} improves all semantic consistency metrics across specificity levels with the largest gain at L1. Also, it widens the Agreement margin over the shuffled baseline from 0.11 to 0.17. Intermediate representation probing for Q2 (\cref{fig:revisit}(b)) shows a consistent trend, with a pronounced increase in L1 Agreement around the middle layers where \ours{} is applied. For asymmetric prompting in Q3 (\cref{fig:revisit}(c)), \ours{} improves the follow rates in both directions. Specifically, the follow rates relative to Both-L3 increase from 26\% to 47\% for video following audio and from 70\% to 75\% for audio following video. The slight increases observed in the Both-L1 and Both-L3 conditions stem from stronger overall semantic representations induced by unimodal alignment. Together, these results demonstrate that \ours{} effectively promotes genuine cross-modal semantic coordination.

\section{Conclusion}
In this work, we rethink joint audio-video generation networks through the lens of \textit{intrinsic jointness}, to look beyond the seemingly strong audio--visual consistency reported under standard evaluations. Using semantically underspecified prompts which leave multiple valid events possible, we show that recent T2AV models largely fail to coordinate audio and video toward a shared semantic realization. To address this, we propose \textit{JointREPA}, a lightweight post-training method that aligns internal representations both within and across modalities. In addition, to rigorously evaluate semantic jointness, we introduce \textit{UNSAIDBench}, which---unlike existing benchmarks with detailed prompts---abstracts fine-grained event identities into general categories. Applying \ours{} to two models with distinct architectures and model sizes yields consistent improvements in semantic consistency across three benchmarks, including \savis{}. Although our study of intrinsic jointness focuses on explicit semantic identities, extending it to physical attributes (e.g., material and weight) offers a promising avenue for physically-coherent T2AV generation. Looking ahead, we envision \textit{intrinsic jointness} as a core design principle for T2AV models, moving beyond mere co-generation toward truly unified audio--video synthesis.
\newpage
{
    \small
    \bibliographystyle{ieeenat_fullname}
    \bibliography{main}
}
\newpage

\end{document}